\documentclass{article}

\usepackage{arxiv}

\usepackage[utf8]{inputenc} 
\usepackage[T1]{fontenc}    
\usepackage{hyperref}       
\usepackage{url}            
\usepackage{booktabs}       
\usepackage{amsfonts}       
\usepackage{nicefrac}       
\usepackage{microtype}      
\usepackage{lipsum}
\usepackage{graphicx}
\usepackage{comment}
\usepackage{multirow}
\usepackage{adjustbox}

\usepackage{subcaption}
\usepackage{amsmath}
\usepackage[psamsfonts]{amssymb}
\usepackage{amsxtra}
\usepackage{threeparttable}

\title{Design of Supervision-Scalable Learning Systems: Methodology and Performance Benchmarking}

\author{
  Yijing Yang \\
  University of Southern California\\
  Los Angeles, California, USA \\
  \texttt{yijingya@usc.edu} \\
     \And
  Hongyu Fu \\
  University of Southern California\\
  Los Angeles, California, USA \\
  \texttt{hongyufu@usc.edu} \\
   \And
  C.-C. Jay Kuo \\
  University of Southern California\\
  Los Angeles, California, USA \\
  \texttt{cckuo@sipi.usc.edu} \\
}

\begin{document}
\maketitle

\begin{abstract}

The design of robust learning systems that offer stable performance
under a wide range of supervision degrees is investigated in this work.
We choose the image classification problem as an illustrative example
and focus on the design of modularized systems that consist of three
learning modules: representation learning, feature learning and decision
learning.  We discuss ways to adjust each module so that the design is
robust with respect to different training sample numbers. Based on these
ideas, we propose two families of learning systems. One adopts the
classical histogram of oriented gradients (HOG) features while the other uses
successive-subspace-learning (SSL) features. We test their performance against
LeNet-5, which is an end-to-end optimized neural network, for MNIST and
Fashion-MNIST datasets. The number of training samples per image class
goes from the extremely weak supervision condition (i.e., 1 labeled
sample per class) to the strong supervision condition (i.e., 4096 labeled
sample per class) with gradual transition in between (i.e., $2^n$, $n=0,
1, \cdots, 12$).  Experimental results show that the two families of
modularized learning systems have more robust performance than LeNet-5.
They both outperform LeNet-5 by a large margin for small $n$ and have
performance comparable with that of LeNet-5 for large $n$.

\end{abstract}

\section{Introduction}\label{sec:introduction}

Supervised learning is the main stream in pattern recognition, computer
vision and natural language processing nowadays due to the great success
of deep learning. On one hand, the performance of a learning system
should improve as the number of training samples increases. On the other
hand, some learning systems may benefit more than others from a large
number of training samples. For example, deep neural networks (DNNs)
often work better than classical learning systems that contain feature
extraction and classification two stages. How the quantity of labeled
samples affects the performance of learning systems is an important
question in the data-driven era. Is it possible to design a
supervision-scalable learning system? We attempt to shed light on these
questions by choosing the image classification problem as an
illustrative example in this work. 

Strong supervision is costly in practice since data labeling demands a
lot of time and resource. Besides, it is unlikely to collect and label
desired training samples in all possible scenarios. Even with a huge
amount of labeled data in place, it may still be substantially less than
the need. Weak supervision can appear in different forms, e.g., inexact
supervision, inaccurate supervision, and incomplete supervision.  Labels
are provided at the coarse grain (instead of the instance level) in
inexact supervision. One example is multi-instance learning
\cite{foulds2010review, wei2016scalable}. For inaccurate supervision,
labels provided may suffer from labeling errors, leading to the noisy
label problem in supervised learning \cite{angluin1988learning,
frenay2013classification}. Only a limited number of labeled data is
available to the training process in incomplete supervision
\cite{zhou2018brief,zhang2021learning}. Here, we consider the scenario
of incomplete supervision. 

To improve learning performance under incomplete supervision, solutions
such as semi-supervised learning and active learning have been
developed.  In semi-supervised learning, both labeled and unlabeled data
are utilized to achieve better performance \cite{chapelle2009semi,
zhu2009introduction,zhou2010semi}.  It is built upon several assumptions
such as smoothness, low-density, and manifold assumptions
\cite{van2020survey}.  In active learning, it attempts to expand the
labeled data set by identifying important unlabeled instances that help
boost the learning performance most
~\cite{settles2009active,haussmann2020scalable}.  Another related
technology is few-shot learning (FSL) \cite{fink2004object} that learns
from a very limited number of labeled data without the help of unlabeled
data. For example, a \textit{N}-way-\textit{K}-shot classification
refers to a labeled set with \textit{K} samples from each of the
\textit{N} classes. Meta learning is often used to solve the FSL problem
\cite{sun2019meta,chen2021meta}. 

Humans can learn effectively in a weakly supervised setting.  In
contrast, deep learning networks often need more labeled data to achieve
good performance. What makes weak supervision and strong supervision
different?  There is little study on the design of supervision-scalable
leaning systems. In this work, we show the design of two learning
systems that demonstrate an excellent scalable performance with respect
to various supervision degrees.  The first one adopts the classical
histogram of oriented gradients (HOG) \cite{dalal2005histograms}
features while the second one uses successive-subspace-learning (SSL)
features.  We discuss ways to adjust each module so that their design is
more robust against the number of training samples.  To illustrate their
robust performance, we compare with the performance of LeNet-5, which is
an end-to-end optimized neural network, for MNIST and Fashion-MNIST
datasets. The number of training samples per image class goes from the
extremely weak supervision condition (i.e., 1 labeled sample per class)
to the strong supervision condition (i.e., 4096 labeled sample per
class) with gradual transition in between (i.e., $2^n$, $n=0, 1, \cdots,
12$).  Experimental results show that the two families of modularized
learning systems have more robust performance than LeNet-5.  They both
outperform LeNet-5 by a large margin for small $n$ and have performance
comparable with that of LeNet-5 for large $n$. 

The rest of the paper is organized as follows.  The design of HOG-based
learning systems is examined in Sec.~\ref{sec:method-HOG}, where two
methods, called HOG-I and HOG-II, are proposed. The design of SSL-based
learning systems is investigated in Sec.~\ref{sec:method}, where two
methods, called IPHop-I and IPHOP-II, are presented. Performance
benchmarking of HOG-I, HOG-II, IPHop-I, IPHop-II and LeNet-5 is conducted
in Sec.~\ref{sec:experiments}. Discussion on experimental results
is given in Sec.~\ref{sec:discussion}.  Finally, concluding remarks and
future work are given in Sec.~\ref{sec:conclusion}. 

\begin{figure*}[tbp]
\centerline{\includegraphics[width=1.0\linewidth]{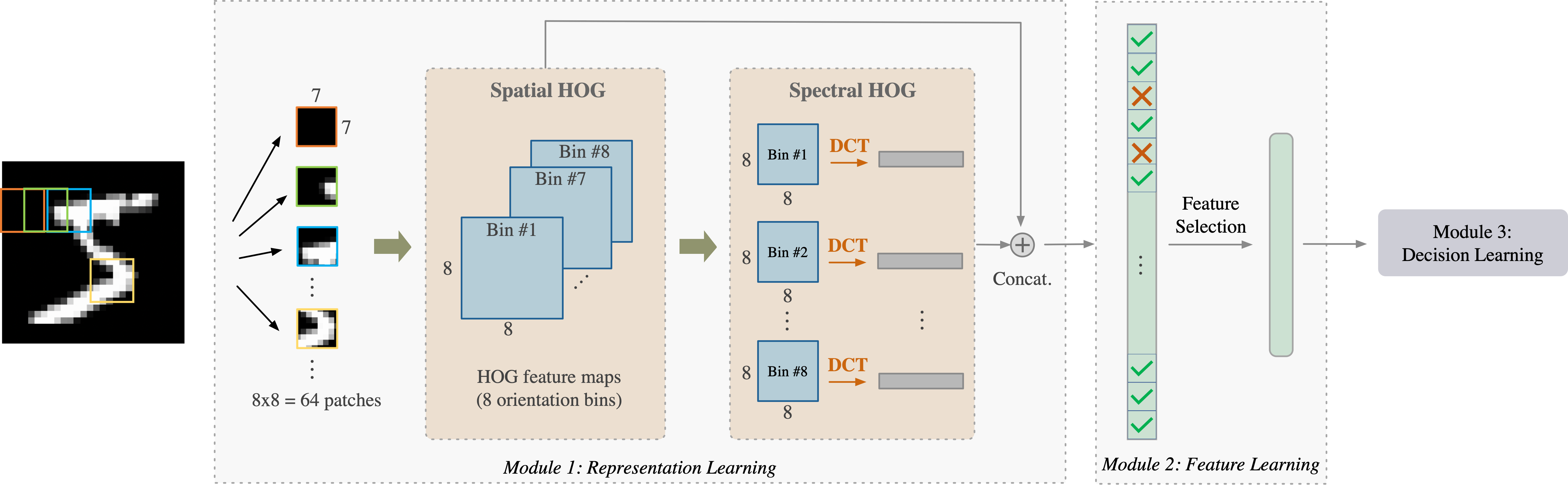}}
\caption{An overview of the HOG-based learning system, where the input image
is of size $28 \times 28$.}\label{fig:hog_pipe}
\end{figure*}
\section{Design of Learning Systems with HOG Features}\label{sec:method-HOG}

Classical pattern recognition methods consist of feature extraction
and classification two steps. One well known feature extraction method
is the Histogram of Oriented Gradients (HOG)
\cite{dalal2005histograms}.  Before the big data era, most datasets are
small in terms of the numbers of training samples and test samples. As a
result, HOG-based solutions are typically applied to small datasets. To
make HOG-based solutions scalable to larger datasets, some modifications
have to be made. In this section, we propose two HOG-based learning
systems, HOG-I and HOG-II. They are suitable for small and large
training sizes, respectively. 

\subsection{Design of Three Modular Components}\label{sec:3_modules}

As mentioned earlier, we focus on the design of a modularized system that
can be decomposed into representation learning, feature learning and
decision learning three modules. We will examine them one by one below. 

\subsubsection{Representation Learning}\label{sec:hog_module1}

HOG was originally proposed for human detection in
\cite{dalal2005histograms}. It measures the oriented gradient
distribution in different orientation bins evenly spaced over 360
degrees at each local region of the input image. Modifications are made
to make the HOG representation more powerful for multi-class
recognition. Images in the MNIST and the Fashion-MNIST datasets have
resolution of $28\times 28$ without padding.  The proposed HOG
representation scheme for them is illustrated in
Fig.~\ref{fig:hog_pipe}. As shown in the figure, both spatial and
spectral HOG representations are considered. Hyper-parameters used in
the experiments are specified below. 

First, we decompose an input image into $7\times 7$ overlapping patches
with stride $3$, leading to $8 \times 8=64$ patches. HOG is computed
within each patch and the number of orientation bins is set to $8$. For
each orientation bin, there are $8\times 8$ responses in the spatial
domain. Thus, each image has a 512-D spatial HOG representation vector.
It is called the spatial HOG representation since each element in the
vector captures the probability of a certain oriented gradient in a
local region.  Next, for each bin, we apply the 2D discrete cosine
transform (DCT) to $8 \times 8$ spatial responses to derive the spectral
representation. The DCT converts 64 spatial responses to 64 spectral
responses. It is called the spectral HOG representation.  Each image has
a 512-D spectral HOG representation vector, too.  We combine spatial and
spectral HOG representations to yield a vector of 1024 dimensions as the
joint spatial/spectral HOG features. 

\subsubsection{Feature Learning}\label{sec:hog_module2}

The size of HOG feature set from Module 1 is large. It is desired to
select discriminant features to reduce the feature dimension before
classification. We adopt two feature selection methods in Module 2 as
elaborated below. 

When the training size is small, we may consider unsupervised feature
selection. One common method is to use the variance of a feature.
Intuitively speaking, if one feature has a smaller variance value among
all training samples, it is not able to separate different classes well
as compared with features that have higher variance values.  Thus, we
can rank order features from the largest to the smallest variance values
and use a threshold to select those of larger variance. 

When the training size becomes larger, we can exploit class labels for
better feature selection.  The advantage of semi-supervised feature
selection over unsupervised becomes more obvious as the supervision
level increases. Here, we adopt a newly developed method, called
Discriminant Feature Test (DFT) \cite{yang2022supervised}, for
semi-supervised feature selection.  DFT computes the discriminant power
of each 1D feature by partitioning its range into two non-overlapping
intervals and searching for the optimal partitioning point that
minimizes the weighted entropy loss.  Mathematically, we have the
entropy function of the left interval as
\begin{equation}\label{eq:entropy}
H^{i}_{L,t} = -\sum_{c=1}^{C}p^{i}_{L,c}log(p^{i}_{L,c}),
\end{equation}
where $p^{i}_{L,c}$ is the probability of class $c$ in the left interval
of the $i$th feature and $t$ is a threshold.  Similarly, we can compute
entropy $H^{i}_{R,t}$ for the right interval.  Then, the entropy of the
whole range is the weighted average of $H_{L,t}$ and $H_{R,t}$, denoted
by $H^i_{t}$. Then, the optimized entropy $H^i_{op}$ for the $i$th feature 
is given by
\begin{equation}\label{eq:optimized_entropy}
H^i_{op} = \min_{t \epsilon T} H^i_{t},
\end{equation}
where $T$ indicates a set of discrete partition points.  The lower the
weighted entropy, the higher the discriminant power. Top $K$ features
with the lowest DFT loss are selected as discriminant features. In
our experiments, we select $K=400$ features out of the 1024 joint
spatial/spectral HOG features for MNIST while setting $K=600$ for
Fashion-MNIST. 

\subsubsection{Decision Learning}\label{sec:hog_module3}

We consider two classifiers - the k-nearest-neighbor (KNN) classifier
and the eXtreme Gradient Boosting (XGBoost~\cite{xgb}) classifier. In a
weakly supervised setting with a small number of training samples, the
choice is very limited and the distance-based classifier seems to be a
reasonable choice.  When the training sample becomes larger, we can use
more powerful supervised classifier to yield better classification
performance. The XGBoost classifier is a representative one. 

\begin{figure*}[tbp]
\centering
\begin{subfigure}{0.8\textwidth}
\centering
    \includegraphics[width=0.93\linewidth]{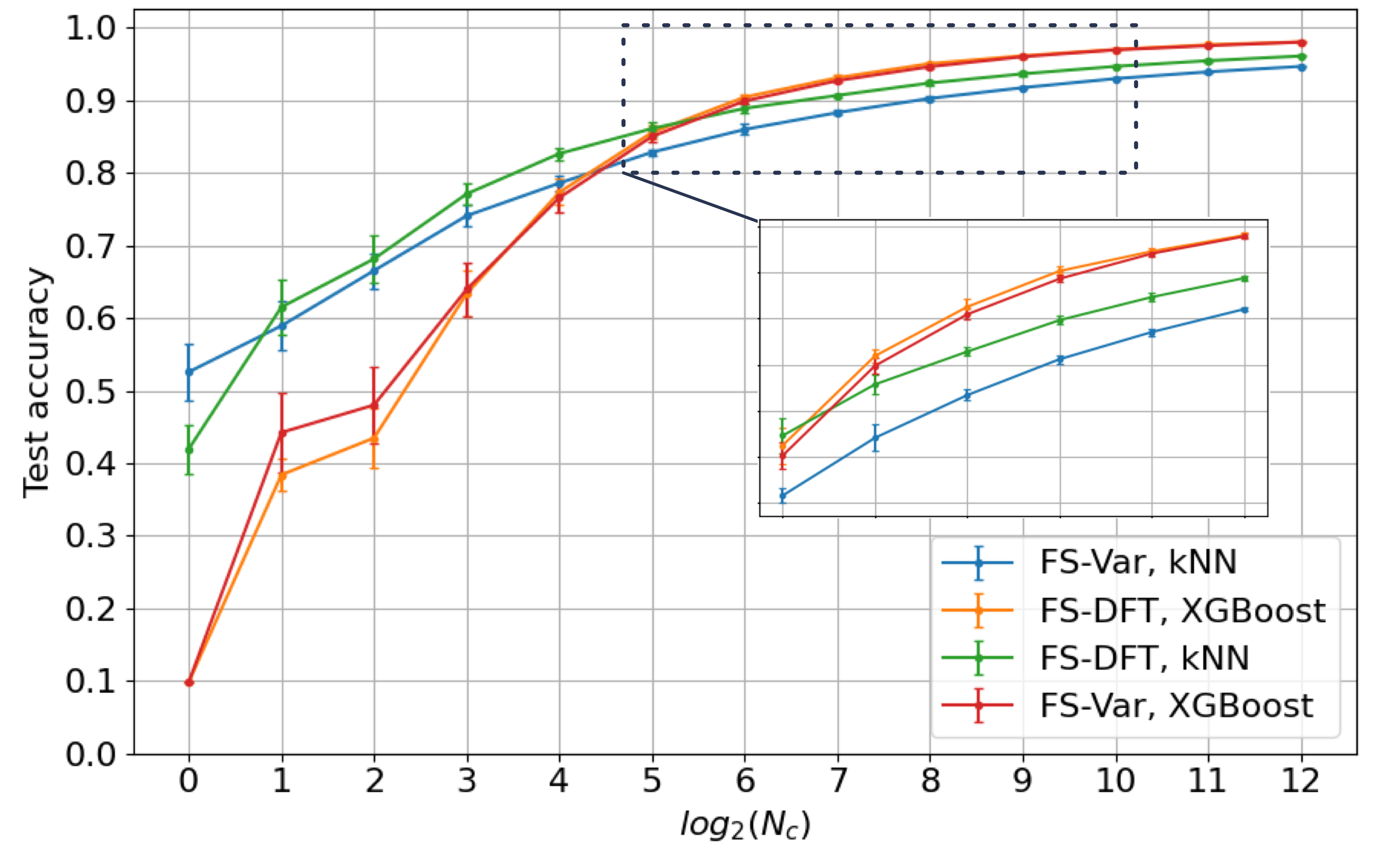}
    \caption{MNIST Dataset}
\end{subfigure}
\begin{subfigure}{0.8\textwidth}
\centering
    \includegraphics[width=0.93\linewidth]{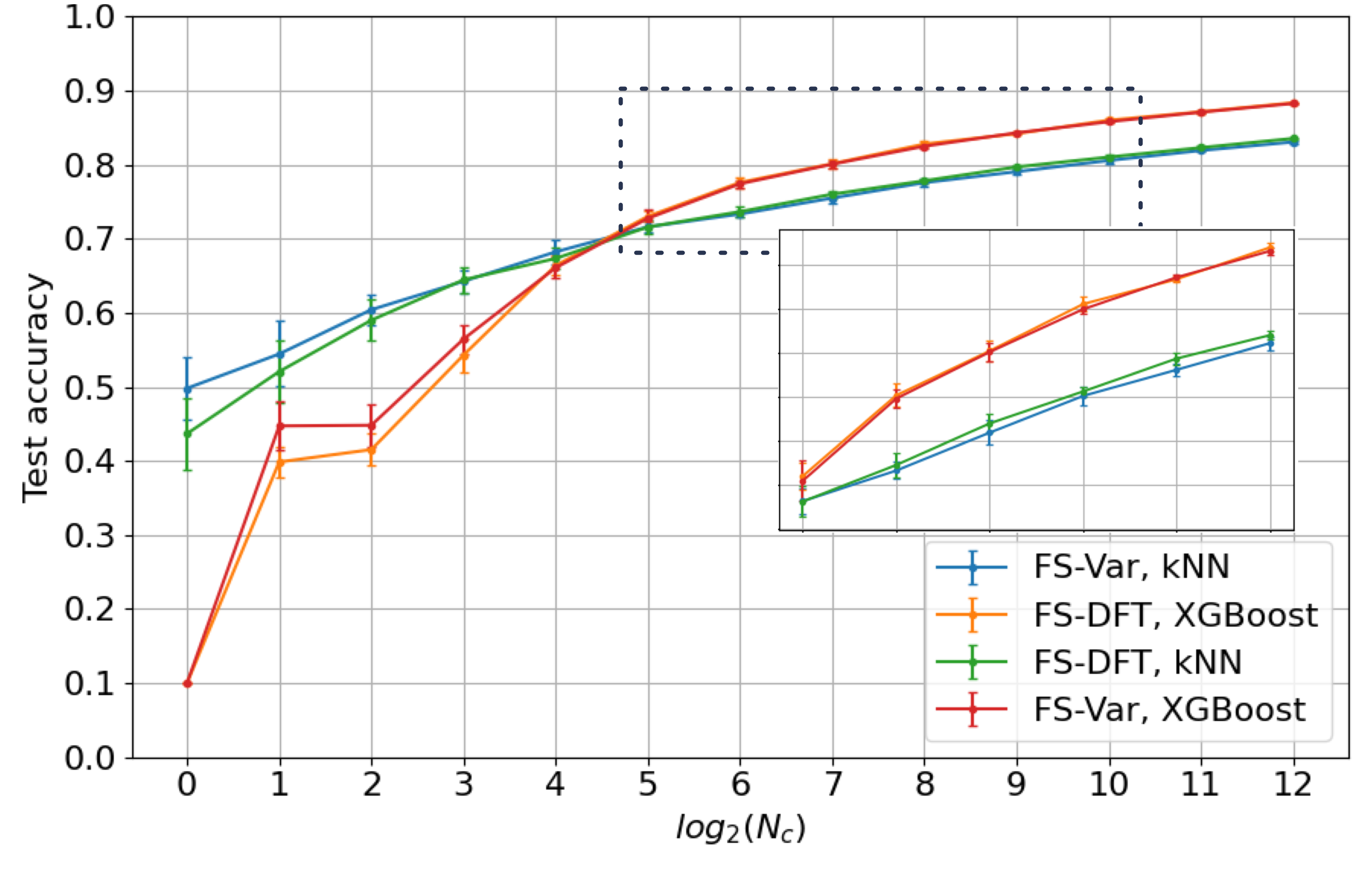}
    \caption{Fashion-MNIST Dataset}
\end{subfigure}
\caption{Performance comparison of HOG-based learning systems on MNIST
and Fashion-MNIST datasets under four different combinations among two
feature learning methods (variance thresholding and DFT) and two
classifiers (KNN and XGBoost) as a function of the training sample
number per class in the log scale.}\label{fig:hog_combination}
\end{figure*}

\subsection{HOG-I and HOG-II}\label{sec:hog_12}

Based on the three modules introduced in Sec. \ref{sec:3_modules}, we
propose two HOG-based learning systems below. 
\begin{enumerate}
\item HOG-I
\begin{itemize}
\item Objective: targeting at weaker supervision
\item Representation Learning: HOG features
\item Feature Learning: variance thresholding
\item Decision Learning: KNN
\end{itemize}
\item HOG-II 
\begin{itemize}
\item Objective: targeting at stronger supervision
\item Representation Learning: HOG features
\item Feature Learning: DFT
\item Decision Learning: XGBoost
\end{itemize}
\end{enumerate}

\begin{figure*}[tbp]
\centering
\begin{subfigure}{0.48\textwidth}
\centering
    \includegraphics[width=1.0\linewidth]{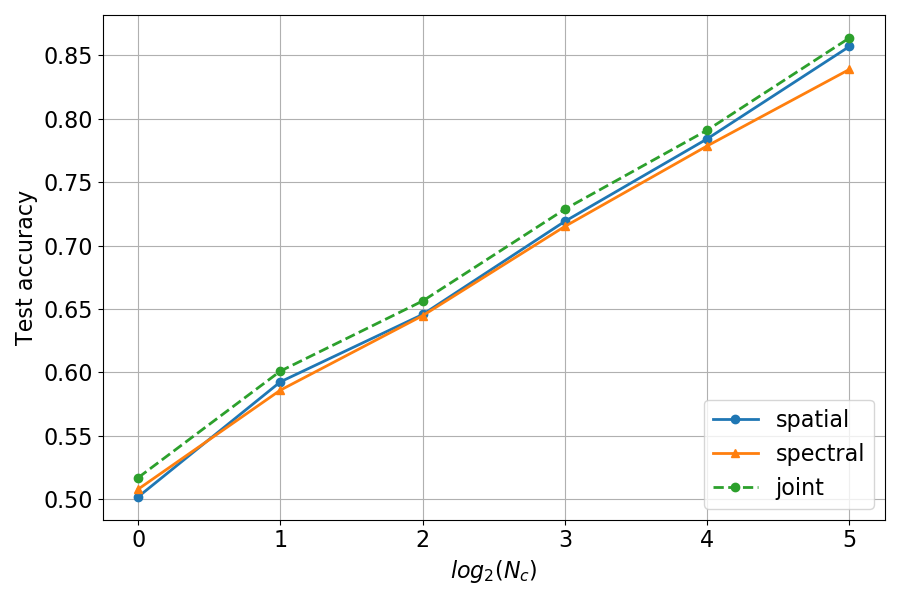}
    \caption{Weak Supervision with HOG-I}
\end{subfigure}
\begin{subfigure}{0.48\textwidth}
\centering
    \includegraphics[width=1.0\linewidth]{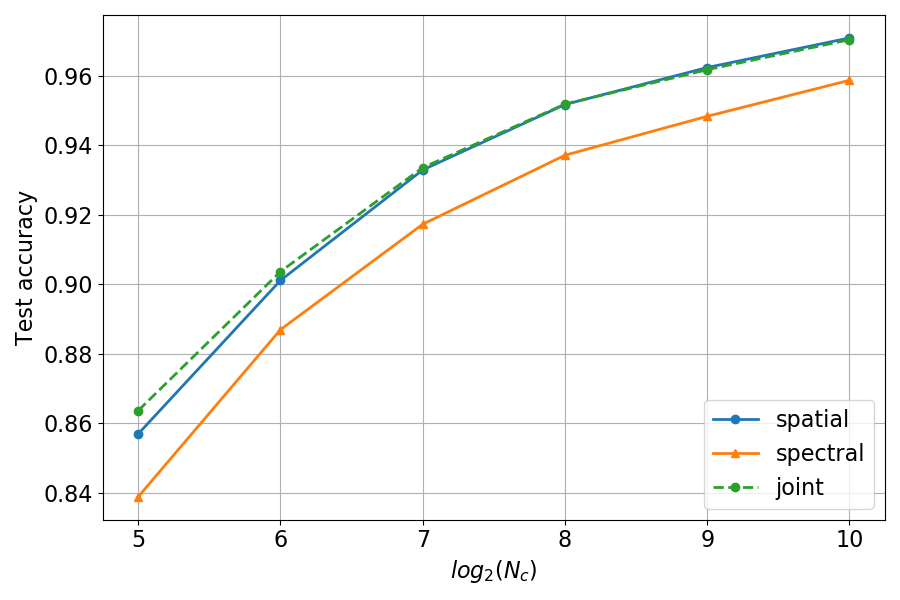}
    \caption{Strong Supervision with HOG-II}
\end{subfigure}
\caption{Performance comparison of spatial, spectral and joint HOG features on 
MNIST under weak and strong supervision conditions.}\label{fig:hog_mnist}
\end{figure*}

\begin{figure*}[tbp]
\centering
\begin{subfigure}{0.48\textwidth}
\centering
    \includegraphics[width=1.0\linewidth]{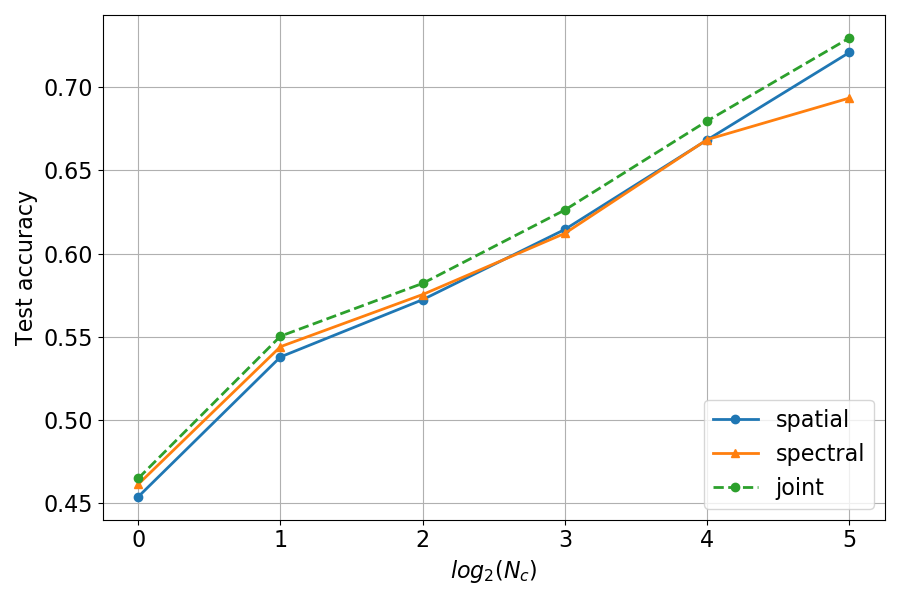}
    \caption{Weak Supervision with HOG-I}
\end{subfigure}
\begin{subfigure}{0.48\textwidth}
\centering
    \includegraphics[width=1.0\linewidth]{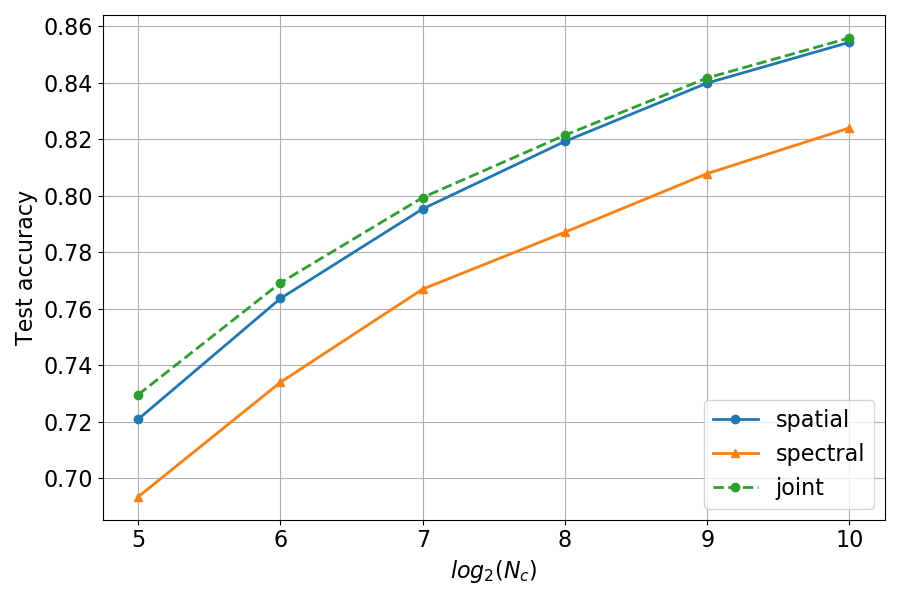}
    \caption{Strong Supervision with HOG-II}
\end{subfigure}
\caption{Performance comparison of spatial, spectral and joint HOG features on 
Fashion-MNIST under weak and strong supervision conditions.}\label{fig:hog_fashion}
\end{figure*}

To justify these two designs, we conduct experiments on MNIST and
Fashion-MNIST to gain more insights. By adopting the joint HOG features,
we consider four combinations of two feature learning methods (variance
thresholding and DFT) and two classifiers (KNN and XGBoost).  The
classification accuracy of test data against MNIST and Fashion-MNIST, as
a function of the number of training data per class, $N_c$, is compared
in Fig. \ref{fig:hog_combination}, whose x-axis is in the unit of
$\log_2 N_c = n_c$. 

The primary factor to the performance is the classifier choice. When
$N_c \le 2^3=8$, the KNN classifier outperforms the XGBoost classifier
by an obvious margin. If $N_c \ge 2^6=64$, the XGBoost classifier offers
better performance. There is a cross over region between $N_c=2^4=16$
and $N_c=2^5=32$.  Furthermore, by zooming into the weak supervision
region with $N_c \le 8$, the variance thresholding feature selection
method is better than the DFT feature selection method. On the other
hand, in the stronger supervision region with $N_c \ge 32$, the DFT
featurer selection method is better than the variance thresholding
feature selection method. For this reason, we use the combination of
variance thresholding and KNN in HOG-I and adopt the combination of DFT
and XGBoost in HOG-II. They target at the weaker and stronger
supervision cases, respectively.

The phenomenon observed in Fig. \ref{fig:hog_combination} can be
explained below. When the supervision degree is weak, it is difficult to
build meaningful data models (e.g., low-dimensional manifolds in a
high-dimensional representation space). Variance thresholding and KNN
are classical feature selection and classification methods derived from
the Euclidean distance, respectively. When the supervision degree
becomes stronger, it is feasible to build more meaningful data models.
The Euclidean distance measure is too simple to capture the data
manifold information. Instead, DFT and XGBoost can leverage the manifold
structure for better feature selection and decision making,
respectively. 

Next, we compare the performance of spatial, spectral and joint HOG
features for HOG-I and HOG-II under their preferred supervision range
for MNIST and Fashion-MNIST. Their results are shown in Figs.
\ref{fig:hog_mnist} and \ref{fig:hog_fashion}, respectively. We have the
following observations.  First, the performance gap between spatial and
spectral HOG features is small under weak supervision with $2^4=16$
training samples per class. The performance gap becomes larger if the
training sample number per class is greater or equal to $2^5=32$.
Second, the joint HOG features provide the best overall performance.
This is not a surprise since the set of joint HOG features contain the
spatial and spectral HOG features as its two subsets. Here, we would
like to point out that the performance gap between the joint HOG
features and the spatial HOG features are larger for smaller $N_c$. It
means that the spectral HOG features do complement the spatial HOG
features and contribute to the performance gain. The value of spectral
HOG features diminishes as $N_c$ is sufficiently large. 

We may give the following explanation to the phenomena observed in Figs.
\ref{fig:hog_mnist} and \ref{fig:hog_fashion}. By performing the DCT on
the histogram of each bin over $8 \times 8$ patches, the values of
low-frequency DCT coefficients are larger due to energy compaction.
Their values for the same object class are relatively stable regardless
of the supervision degree. In contrast, the spatial HOG features are
distributed over the whole image. They are more sensitive to the local
variation of each individual sample. Thus, when the supervision is weak,
HOG-I can benefit from spectral HOG features. Second, as the supervision
becomes stronger, the situation is different. Although the HOG of a
single patch provides only the local information, we can obtain both
local and global information by concatenating spatial HOG features
across all patches. The HOG of at the same patch location could be noisy
(i.e. varying from one sample to the other). Yet, the variation can be
filtered out by DFT and XGBoost. On the other hand, the values of
high-frequency DCT coefficients are small and many of them are close to
zero because of energy compaction. Thus, spectral HOG features are not
as discriminant as spatial HOG features under strong supervision.

\begin{figure*}[tbp]
\centerline{\includegraphics[width=1.0\linewidth]{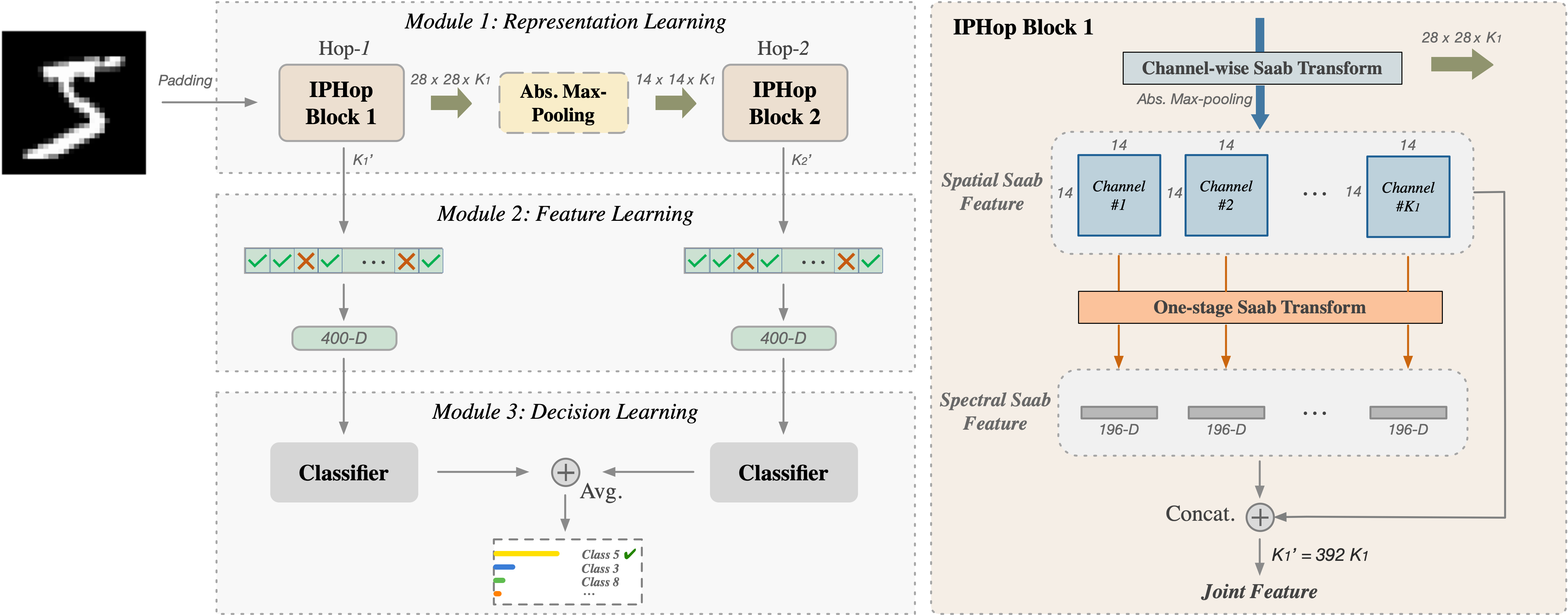}}
\caption{An overview of the SSL-based learning system, where the input image
is of size $32 \times 32$.}\label{fig:pipeline}
\end{figure*}
\section{Design of Learning Systems with SSL Features}\label{sec:method}

Successive subspace learning (SSL) was recently introduced in
\cite{kuo2016understanding, kuo2017cnn,
kuo2018data,kuo2019interpretable}.  The technique has been applied to
many applications such as point cloud classification, segmentation and
registration \cite{zhang2020pointhop, zhang2020pointhop++,
zhang2020unsupervised, kadam2020unsupervised, kadam2022r}, face
recognition \cite{rouhsedaghat2020facehop, rouhsedaghat2021successive},
deepfake detection \cite{chen2021defakehop}, anomaly detection
\cite{zhang2021anomalyhop}, etc. SSL-based object classification work
can be found in~\cite{chen2020pixelhop,
chen2020pixelhop++,Yang2021EPixelHopAE}.  We propose two improved
PixelHop (IPHop) learning systems and name them IPHop-I and IPHop-II in
this section. 

The system diagram of IPHop-I/II is shown in the left subfigure of
Fig.~\ref{fig:pipeline}.  It consists of three modules: 1) unsupervised
representation learning based on SSL features, 2) semi-supervised
feature learning, and 3) supervised decision learning.  Since its
modules 2 and 3 are basically the same as those in HOG-based learning
systems, we will primarily focus on the representation learning in Sec.
\ref{sec:iphop_module1}. Afterwards, we compare the performance of
IPHop-I and IPHop-II under weak and strong supervision scenarios in
Sec.~\ref{sec:iphop2}. 

\subsection{SSL-based Representation Learning}\label{sec:iphop_module1}

We describe the processing procedure in Module 1 of the left subfigure
of Fig.~\ref{fig:pipeline} below.  The input is a tiny image of spatial
resolution $32 \times 32$.  The processing procedure can be decomposed
into two cascaded units, called Hop-1 and Hop-2, respectively.  We first
extract the spatial Saab features at Hop-1 and Hop-2.  For each hop
unit, we apply filters of spatial size $5\times5$.  At Hop-1, a
neighborhood of size $5\times5$ centered at each of the interior
$28\times28$ pixels is constructed. The Saab transform is conducted at
each neighborhood to yield $K_1=25$ channel responses at each pixel.
Afterwards, a $2\times2$ absolute max-pooling is applied to each
channel. It reduces the spatial resolution from $28\times28$ to
$14\times14$. As a result, the input to Hop-2 is $14 \times 14 \times
25$.  Similarly, we apply the channel-wise Saab transform with $K_2$
filters to the interior $10\times10$ points to get $K_2$ responses for
each point. Here, we set $K_2=256$ and $K_2=204$ for MNIST and
Fashion-MNIST, respectively, based on the energy thresholding criterion
introduced in \cite{chen2020pixelhop++}. 

This above design is basically the standard PixelHop++ pipeline as
described in \cite{chen2020pixelhop++}. The only modification in IPHop
is that we change max-pooling in PixelHop++ to absolute max-pooling.
Note that The responses from Hop-1 can be either positive or negative
since no nonlinear activation is implemented at Hop-1.  Instead of
clipping negative values to zero, we find that it is advantageous to
take the absolute value of the response first and then conduct the
maximum pooling operation. 

The spatial filter responses extracted at Hop-1 and Hop-2 only have a
local view on the object due to the limited receptive field. They are
not discriminant enough for semantic-level understanding. Since there
exists correlations among these local filter responses, we can conduct
another Saab transform across all local responses at each individual
channel. Such a processing step provides the global spectral Saab
features at Hop-1 and Hop-2 as shown in the right subfigure of
Fig.~\ref{fig:pipeline}. To explain the procedure in detail, we use
Hop-1 as an example. For each of the $K_1=25$ channels, $14 \times
14=196$ spatial Saab features are flattened and then passed through a
one-stage Saab transform.  All responses are kept without truncation.
Thus, the dimension of the output spectral Saab features is 196 for each
channel.  As compared to features learned by gradually enlarging the
neighborhood range, the spectral Saab features capture the long range
information from a finer scale. Finally, the spatial and spectral Saab
features are concatenated at Hop-1 and Hop-2 to form the
joint-spatial-spectral Saab features. 

\subsection{IPHop-I and IPHop-II}\label{sec:iphop2}

Since the two hops have different combinations of spatial and spectral
information, it is desired to treat them differently. For this reason, 
we partition IPHop features into two sets:
\begin{itemize}
\item Feature Set no. 1: spatial and spectral features of Hop-1,
\item Feature Set no. 2: spatial and spectral features of Hop-2.
\end{itemize}
Feature learning is used to select the subset of discriminant features
from the raw representation. By following HOG-I and HOG-II, we consider
variance thresholding and DFT two choices, apply them to feature sets
no. 1 and no. 2, and select the same number of optimal features from
each set individually. Furthermore, the same two classifiers are used
for decision learning: KNN and XGBoost.  For KNN, we concatenate optimal
features from feature set no. 1 and no. 2, and compute the distance in this
joint feature space. For XGBoost, we apply it to feature set no. 1 and
no. 2 and make soft decision for each hop separately.  Afterwards, we
average the two soft decisions and use the maximum likelihood principle
to yield the final decision. 

We propose two SSL-based leanring systems below. 
\begin{enumerate}
\item IPHop-I
\begin{itemize}
\item Objective: targeting at weaker supervision
\item Representation Learning: Joint SSL features (i.e. both feature set nos. 1 and 2)
\item Feature Learning: variance thresholding
\item Decision Learning: KNN
\end{itemize}
\item IPHop-II 
\begin{itemize}
\item Objective: targeting at stronger supervision
\item Representation Learning: Joint SSL features (i.e. both feature set nos. 1 and 2)
\item Feature Learning: DFT
\item Decision Learning: XGBoost
\end{itemize}
\end{enumerate}

\begin{figure*}[tbp]
\centering
\begin{subfigure}{0.8\textwidth}
\centering
    \includegraphics[width=0.93\linewidth]{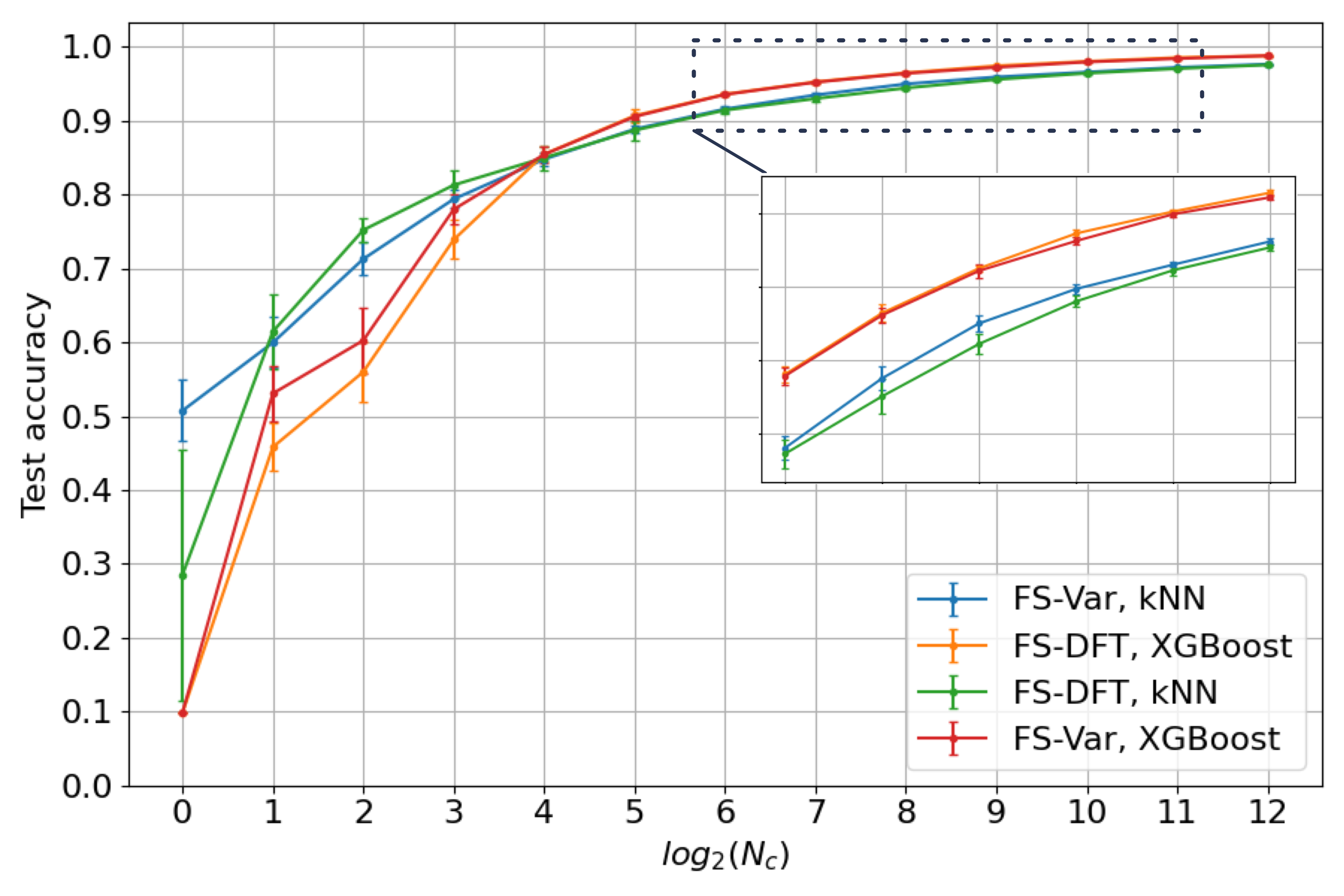}
    \caption{MNIST Dataset}
\end{subfigure}
\begin{subfigure}{0.8\textwidth}
\centering
    \includegraphics[width=0.93\linewidth]{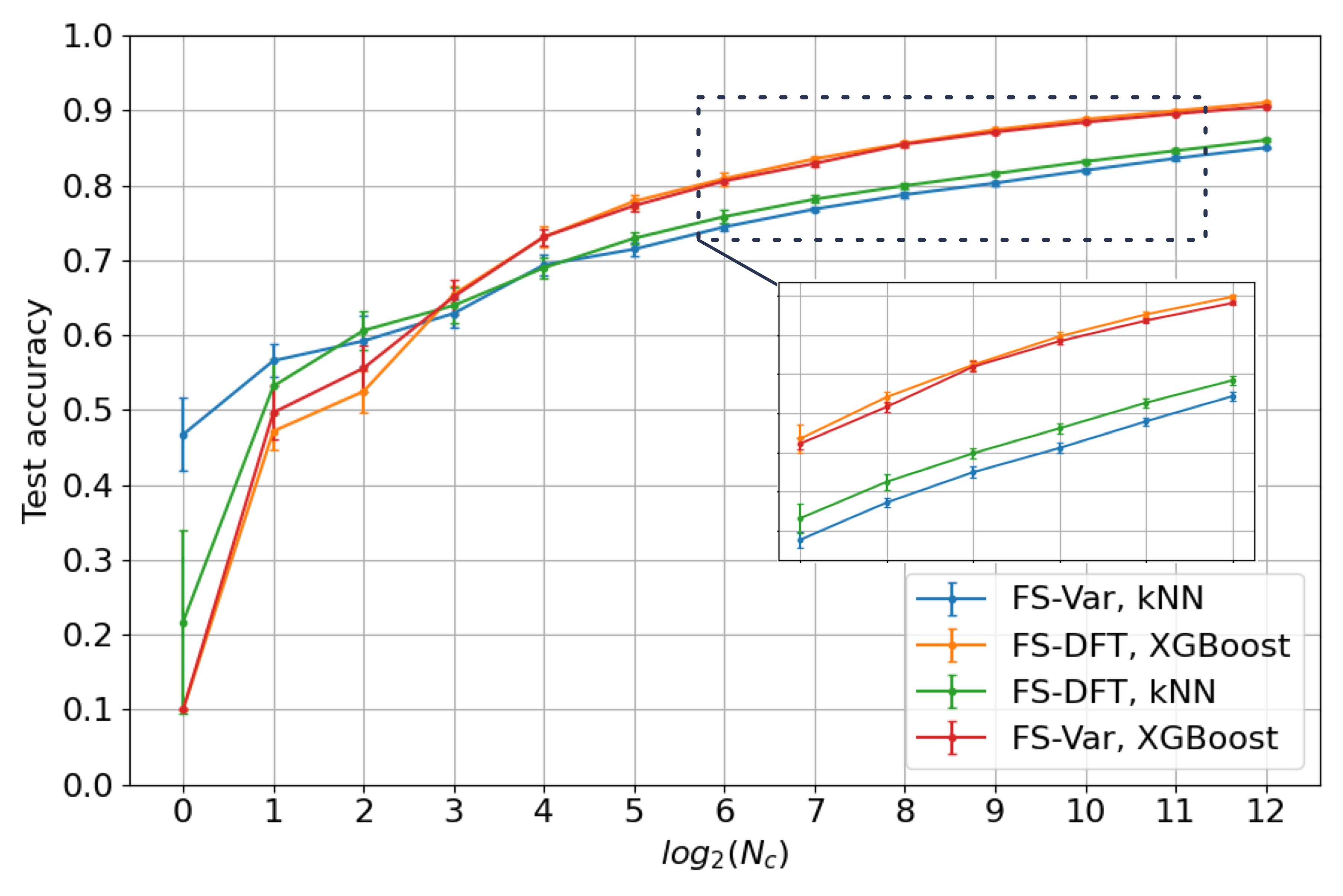}
    \caption{Fashion-MNIST Dataset}
\end{subfigure}
\caption{Performance comparison of SSL-based learning systems on MNIST
and Fashion-MNIST datasets under four different combinations among two
feature learning methods (variance thresholding and DFT) and two
classifiers (KNN and XGBoost) as a function of $n_c=\log_2(N_c)$.}
\label{fig:iphop_combination}
\end{figure*}

To justify these two designs, we consider all four possible combinations
of feature and decision learning choices and compare their performance
in Fig.  \ref{fig:iphop_combination}. We use the fashion-MNIST dataset
as an example in the following discussion. We see from Fig.
\ref{fig:iphop_combination}(b) that KNN outperforms XGBoost under weak
supervision (i.e. the training sample number per class $N_c \le 4$).  On
the other hand, XGBoost outperforms KNN under stronger supervision ($N_c
\ge 16$). There is a transition point at $N_c=8$. For the weak
supervision scenario, variance thresholding feature selection is better
than DFT. This is particularly obvious when $N_c=1$. The performance gap
is around 25\%.  Thus, we use the combination of variance thresholding
and KNN in IPHop-I to be used for weaker supervision.  For the stronger
supervision case, DFT is slightly better than variance thresholding in
both KNN and XGBoost. Therefore, we use the combination of DFT and
XGBoost in IPHop-II to be used for stronger supervision. 

\begin{figure*}[tbp]
\centering
\begin{subfigure}{0.48\textwidth}
\centering
    \includegraphics[width=1.0\linewidth]{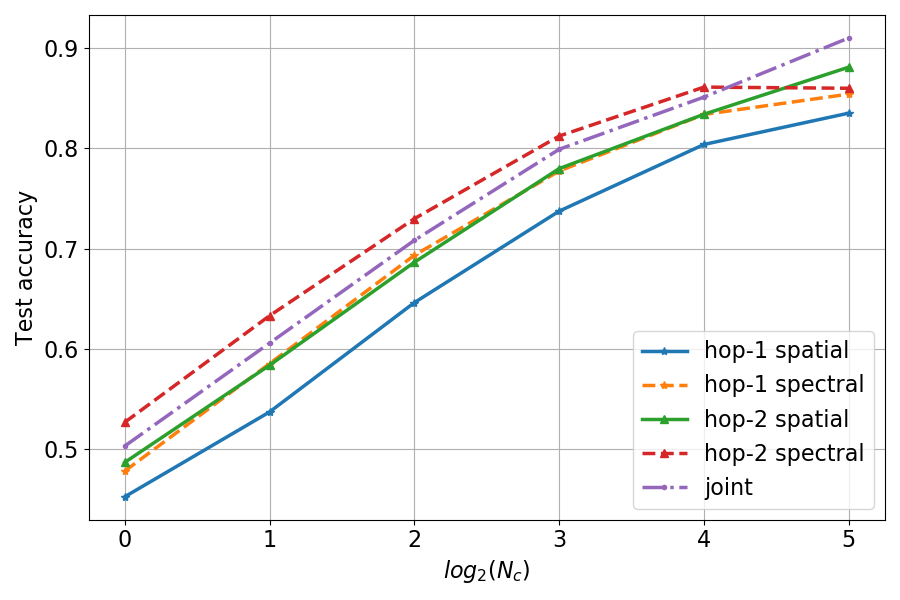}
    \caption{Weak Supervision with IPHop-I}
\end{subfigure}
\begin{subfigure}{0.48\textwidth}
\centering
    \includegraphics[width=1.0\linewidth]{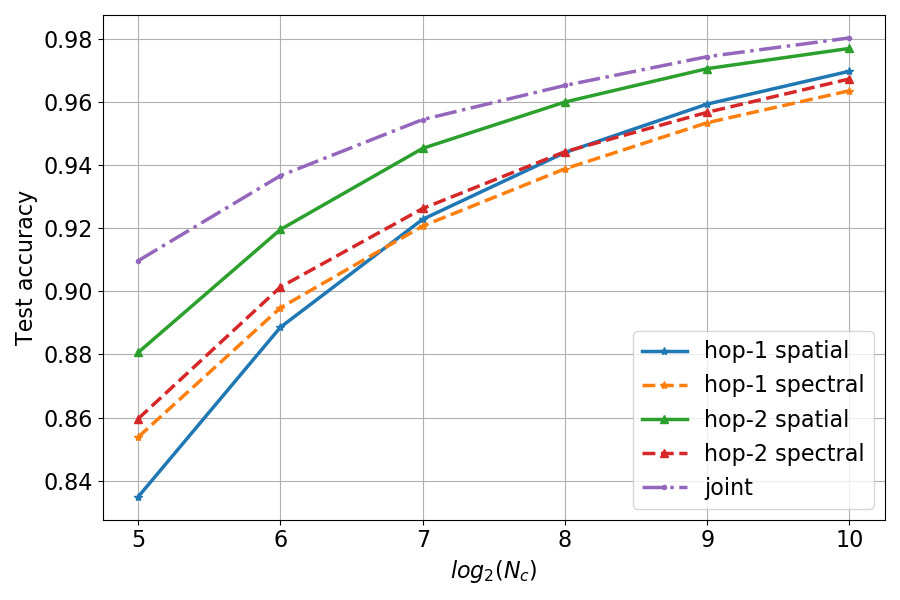}
    \caption{Strong Supervision with IPHop-II}
\end{subfigure}
\caption{Performance comparison of spatial, spectral and joint SSL features on 
MNIST under strong and weak supervision conditions.}\label{fig:iphop_mnist}
\end{figure*}

Next, we conduct ablation study on different representations for IPHop-I
and IPHop-II in their preferred operating ranges to understand the
impact of each feature type.  Fig. \ref{fig:iphop_mnist} compares the
test accuracy with individual spatial and spectral features of hop-1 and
hop-2 and jointly for MNIST under different supervision levels.  

Under weak supervision, we see from Fig. \ref{fig:iphop_mnist}(a) that
spectral features are more powerful than spatial features while spectral
features of hop-2 are slightly better than those of hop-1. Under
stronger supervision, we see from Fig.  \ref{fig:iphop_mnist}(b) that
spatial features are more powerful than spectral features since spatial
features can capture more detail information without energy compaction
and the detail information does help the classification performance as
the number of labeled sample increases. Furthermore, features of hop-2
are more useful than those of hop-1. The main differences between hop-1
and hop-2 features lie two factors:
\begin{itemize}
\item spatial features are determined by the receptive field 
of Saab filters,
\item spectral features are determined by spatial aggregation 
of Saab responses over the entire set of grid points.
\end{itemize}
For the former, the cascaded filters in hop-2 offer a larger receptive
field which has stronger discriminant power than hop-1. For the latter,
hop-1 has 28x28 grid points while hop-2 has only 14x14 grid points.  The
content in hop-1 has larger diversity than that in hop-2. Although the
spatial Saab transform can achieve energy compaction, the percentages of
stable and discriminant spectral features in hop-1 tend to be fewer than
those in hop-2. Yet, hop-1 and hop-2 do provide complementary features
so that the joint feature set gives the best performance. 

\begin{figure*}[tbp]
\centering
\begin{subfigure}{0.48\textwidth}
\centering
    \includegraphics[width=1.0\linewidth]{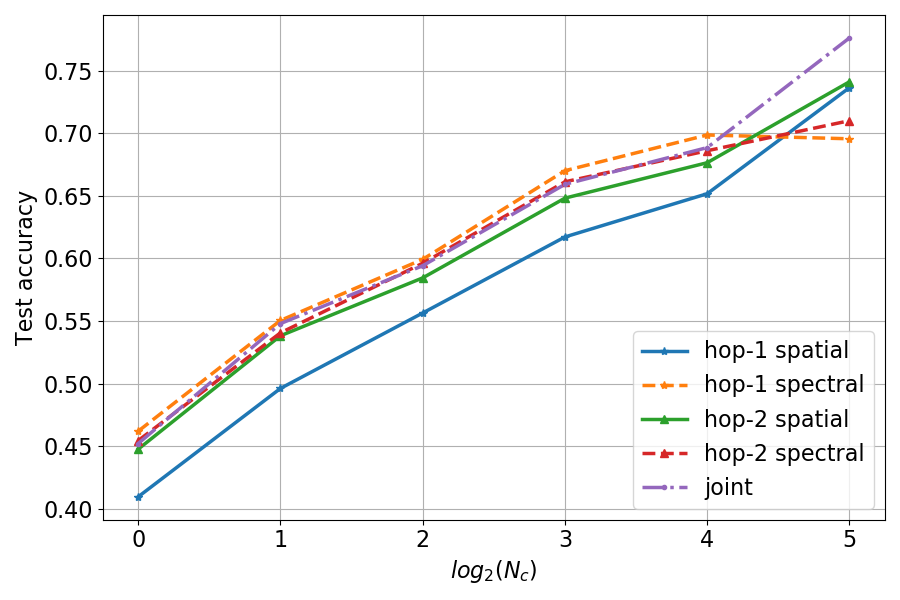}
    \caption{Weak Supervision with IPHop-I}
\end{subfigure}
\begin{subfigure}{0.48\textwidth}
\centering
    \includegraphics[width=1.0\linewidth]{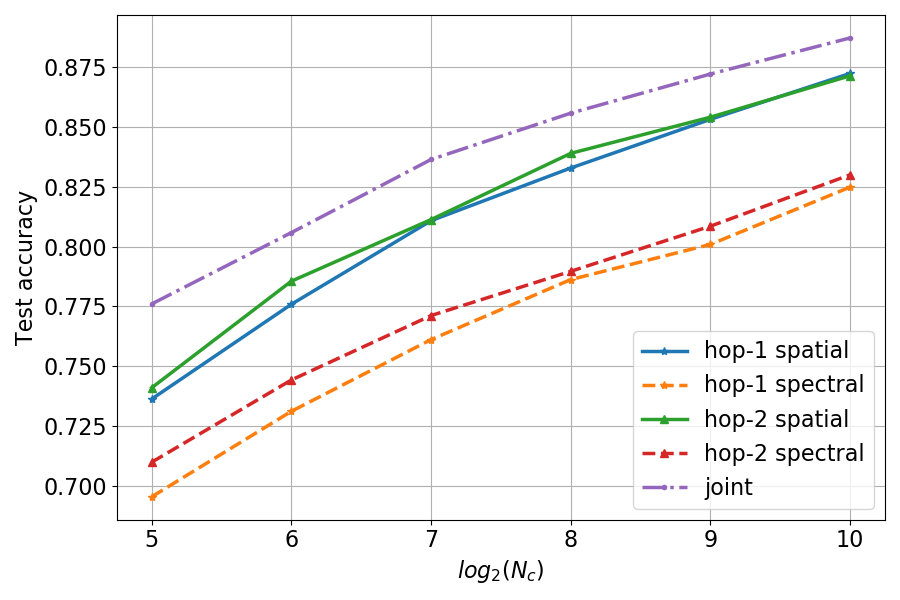}
    \caption{Strong Supervision with IPHop-II}
\end{subfigure}
\caption{Performance comparison of spatial, spectral and joint SSL features on 
Fashion-MNIST under strong and weak supervision conditions.}\label{fig:iphop_fashion}
\end{figure*}

Finally, we show the test accuracy with individual spatial and spectral
features of hop-1 and hop-2 and joint feature sets under different
supervision levels for Fashion-MNIST in Fig. \ref{fig:iphop_fashion}.
The same observations and discussion apply to Fashion-MNIST. 

\section{Experiments}\label{sec:experiments}

Experiments are conducted on MNIST~\cite{lecun1998gradient} and
Fashion-MNIST~\cite{xiao2017fashion} datasets for performance
benchmarking of HOG-I/II, IPHop-I/II and LeNet-5 learning systems
against a wide range of supervision levels.  For HOG and IPHop, we also
introduce a hybrid solution. That is, type I is used when $N_c \le 8$
while type II is used for $N_c \ge 16$. They are called hybrid
HOG and hybrid IPHop, respectively.

\subsection{Experimental Setup}\label{sec:setting}

The Adam optimizer is used for backpropagation in the training of
LeNet-5 network. The number of epochs is set to 50 for all $N_c$ values.
Both MNIST and Fashion-MNIST contain grayscale images of resolution
$28\times28$, with 60K training and 10K test images.  MNIST contains 10
hand-written digits (from 0 to 9) in MNIST while Fashion-MNIST has 10
fashion classes. The training sample number per class is around 6K.
Among the 6K training samples per class, we choose its subset of size
$N_c=2^{n_c}$, $n_c=0, 1, \cdots, 12$ randomly as the training set.  All
classes have the same training sample number.  In words, we go from the
extremely weak supervision condition with 1 labeled sample per class to
the strong supervision condition with 4,096 labeled sample per class
with gradual transition in between.  Experiments with random training
sample selection are performed with multiple runs. The dimension of
representations and features for HOG and IPHop learning systems are
summarized in Table~\ref{tab:HOG_dimension_summary} and
Table~\ref{tab:IPHop_dimension_summary}, respectively. 

\begin{table}[h]
\small
\centering
\caption{Summary of output dimensions of the HOG framework for 
MNIST and Fashion-MNIST.}\label{tab:HOG_dimension_summary}
\begin{tabular}{|cl|cc|}
\hline
\multicolumn{2}{|l|}{}                                                       & MNIST & Fashion-MNIST \\ \hline
\multicolumn{1}{|c|}{\multirow{3}{*}{Module-1}} & Spatial HOG                & 512   & 512           \\
\multicolumn{1}{|c|}{}                          & Spectral HOG               & 512   & 512           \\
\multicolumn{1}{|c|}{}                          & Joint-Spatial-Spectral HOG & 1,024 & 1,024         \\ \hline
\multicolumn{2}{|c|}{Module-2}                                               & 400   & 600           \\ \hline
\end{tabular}
\end{table}

\begin{table}[h]
\small
\centering
\caption{Summary of output resolutions/dimensions of IPHop-II
for MNIST and Fashion-MNIST.}\label{tab:IPHop_dimension_summary}
\begin{tabular}{|cl|cc|cc|}
\hline
\multicolumn{2}{|l|}{\multirow{2}{*}{}} & \multicolumn{2}{c|}{MNIST} & \multicolumn{2}{c|}{Fashion-MNIST} \\ \cline{3-6}
\multicolumn{2}{|l|}{}                  & Hop-1  & Hop-2  & Hop-1  & Hop-2  \\ \hline
\multicolumn{1}{|c|}{\multirow{3}{*}{Module-1}} & Spatial Saab & (14$\times$14)$\times$25   & (5$\times$5)$\times$256   & (14$\times$14)$\times$25       & (5$\times$5)$\times$204       \\
\multicolumn{1}{|c|}{} & Spectral Saab  & 196$\times$25 & 25$\times$256 & 196$\times$25 & 25$\times$204 \\ 
\multicolumn{1}{|c|}{}          & Joint-Spatial-Spectral Saab & 9,800  & 12,800 & 9,800  & 10,200 \\ \hline
\multicolumn{2}{|c|}{Module-2} & 400    & 400    & 400    & 400    \\ \hline
\end{tabular}
\end{table}

\subsection{Performance Benchmarking}\label{sec:benchmarking}

We conduct performance benchmarking of HOG-I, HOG-II, IPHop-I, IPHop-II,
and LeNet-5 in this subsection.  The mean test accuracy and standard
deviation values for MNIST and Fashion-MNIST under different supervision
levels are reported in Table~\ref{tab:mnist_compare} and
Table~\ref{tab:fashion_compare}, respectively, based on results from 10
runs. We have the following observations.
\begin{itemize}
\item Under weak supervision with $N_c=1, \, 2, \, 4, \, 8$, HOG-I and
IPHop-I outperforms LeNet-5 by a large margin. Specifically, where these
is only one labeled image per class ($N_c=1$) or 10 labeled data for the
whole dataset, HOG-I and IPHop-I can still reach an accuracy of around
$50\%$ on for both datasets. For MNIST, HOG-I and IPHop-I surpass
LeNet-5 by 12.51\% and 10.67\%, respectively. For Fashion-MNIST, the
performance gains of HOG-I and IPHop-I are 8.62\% and 5.55\%,
respectively. It shows that the performance of HOG-based and IPHop
learning systems is more robust as the number of labeled samples
decreases. 
\item Under middle supervision with $N_c=16, \, 32, \, 64, \, 128$,
HOG-I, HOG-II, IPHop-I and IPHop-II still outperform LeNet-5.
Furthermore, we start to see the advantage of HOG-II over HOG-I and the
advantage of IPHop-II over IPHop-I.  Besides comparison of mean accuracy
scores, we see that the standard deviation of LeNet-5 is significantly
higher than that of HOG-I, HOG-II, IPHop-I and IPHop-II under the weak
and middle supervision levels. 
\item Under strong supervision with $N_c \ge 1,024$ (or the total
training sample number is more than 10,240 since there are 10 classes), the
advantage of LeNet-5 starts to show up. Yet, IPHop-II still outperforms
LeNet-5 in Fashion-MNIST while the performance difference between
IPHop-II and LeNet-5 on MNIST is very small. 
\item When the full training dataset (i.e., $N_c=6K$) is used, each of
HOG-I, HOG-II, IPHop-I and IPHop-II has a single test accuracy value and
the standard deviation value is zero since the training set is the same.
In contrast, even with the same input, LeNet-5 can yield different
accuracy values due to the stochastic optimization nature of
backpropagation. 
\end{itemize}

\begin{figure*}[tbp]
\centering
\begin{subfigure}{0.8\textwidth}
\centering
    \includegraphics[width=0.93\linewidth]{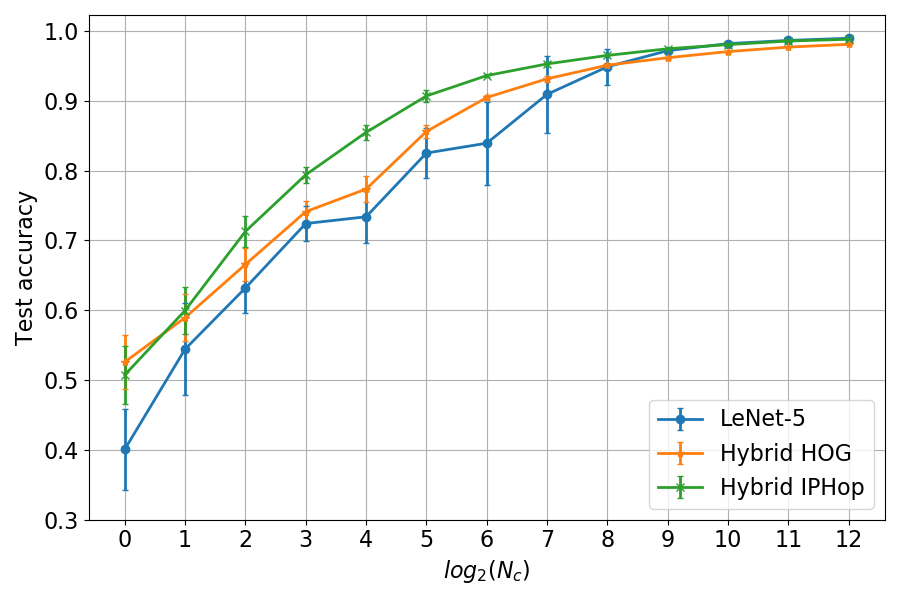}
    \caption{MNIST}
\end{subfigure}
\begin{subfigure}{0.8\textwidth}
\centering
    \includegraphics[width=0.93\linewidth]{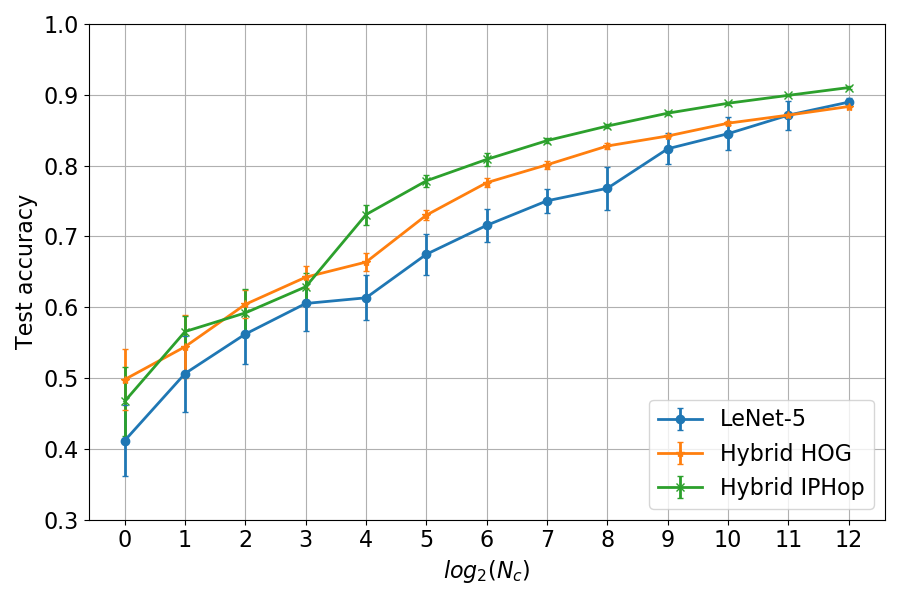}
    \caption{Fashion-MNIST}
\end{subfigure}
\caption{Comparison of test accuracy between hybrid HOG, hybrid IPHop, and
LeNet-5 for MNIST and Fashion-MNIST. For hybrid HOG and IPHop, type I is
adopted when $N_c \le 8$ and type II is adopted for $N_c \ge 16$.}
\label{fig:hybrid_compare}
\end{figure*}

It is natural to consider hybrid HOG and IPHop schemes, where type I is
adopted when $N_c \le 8$ and type II is adopted for $N_c \ge 16$. For
ease of visual comparison, we plot the mean accuracy curves as well as
the standard deviation values (indicated by vertical bars) of hybrid
HOG, hybrid IPHop and LeNet-5 as a function of $N_c$ in Fig.
\ref{fig:hybrid_compare}. Clearly, hybrid IPHop provides the best
overall performance among the three. Hybrid IPHop outperforms LeNet-5 by
a significant margin when $N_c \le 128$ on MNIST and throughout the
whole range of $N_c$ on Fashion-MNIST. As to hybrid HOG, it outperforms
LeNet-5 with $N_c \le 128$, underperforms LeNet-5 with $N_c \ge 512$ and
has a crossover point with LeNet-5 at $N_c=256$ on MNIST. Hybrid HOG has
higher accuracy than LeNet-5 when $N_c \le 1,024$ while its performance
is comparable with LeNet-5 when $N_c = 2,048$ or $4,096$. 

\begin{table*}[tbp]
\small
\centering
\caption{Comparison of the mean test accuracy (\%) and standard deviation 
on MNIST under weak, middle and strong Supervision degree, where the best performance
is highlighted in bold.} \label{tab:mnist_compare}
\begin{tabular}{|c|l|c|cc|cc|} \hline
\multicolumn{1}{|l|}{Supervision} &
  N\_c &
  LeNet-5 &
  HOG-I &
  HOG-II &
  IPHop-I &
  IPHop-II \\ \hline
\multirow{4}{*}{Weak}   & 1    &40.07 ($\pm$5.78) &\textbf{52.58} ($\pm$3.89)   &9.80 ($\pm$0.00) & 50.74 ($\pm$4.13)       & 9.80 ($\pm$0.00)\\
                        & 2    &54.43 ($\pm$6.62) &58.94 ($\pm$3.33)  &38.40 ($\pm$2.14)        & \textbf{59.96} ($\pm$3.42) & 45.82 ($\pm$3.23) \\
                        & 4    &63.19 ($\pm$3.52) &66.55 ($\pm$2.42)  &43.48 ($\pm$4.18)        & \textbf{71.28} ($\pm$2.22) & 56.00 ($\pm$3.98) \\
                        & 8    &72.41 ($\pm$2.50) &74.12 ($\pm$1.50)  &63.39 ($\pm$3.21)        & \textbf{79.40} ($\pm$1.18) & 73.90 ($\pm$2.62) \\ \hline
\multirow{4}{*}{Middle} & 16   &73.38 ($\pm$3.69) &78.61 ($\pm$1.04)  &77.35 ($\pm$1.80)        & 84.78 ($\pm$0.87) & \textbf{85.47} ($\pm$1.02)\\
                        & 32   &82.51 ($\pm$3.62) &82.87 ($\pm$0.40)  &85.60 ($\pm$0.99)        & 88.87 ($\pm$0.44)         & \textbf{90.69} ($\pm$0.85)\\
                        & 64   &83.92 ($\pm$5.94) &86.01 ($\pm$0.72)  &90.47 ($\pm$0.34)        & 91.60 ($\pm$0.32)         & \textbf{93.60} ($\pm$0.21)\\
                        & 128  &90.92 ($\pm$5.52) &88.34 ($\pm$0.30)  &93.14 ($\pm$0.43)        & 93.49 ($\pm$0.32)         & \textbf{95.27} ($\pm$0.24)\\ \hline
\multirow{6}{*}{Strong} & 256  &94.87 ($\pm$2.61) &90.29 ($\pm$0.23)  &95.09 ($\pm$0.26)        & 94.99 ($\pm$0.23)         & \textbf{96.49} ($\pm$0.08)\\
                        & 512  &97.17 ($\pm$0.26) &91.77 ($\pm$0.20)  &96.16 ($\pm$0.15)        & 95.93 ($\pm$0.14)         & \textbf{97.44} ($\pm$0.09)\\
                        & 1024 &\textbf{98.18} ($\pm$0.16) &93.02 ($\pm$0.12)  &97.04 ($\pm$0.14)  & 96.59 ($\pm$0.08)      & 98.04 ($\pm$0.07)\\
                        & 2048 &\textbf{98.64} ($\pm$0.17) &93.95 ($\pm$0.12)  &97.68 ($\pm$0.04)  & 97.23 ($\pm$0.08)      & 98.55 ($\pm$0.07)\\
                        & 4096 &\textbf{98.95} ($\pm$0.09) &94.70 ($\pm$0.13)  &98.08 ($\pm$0.04)  & 97.66 ($\pm$0.06)      & 98.90 ($\pm$0.06)\\
                        & Full &\textbf{99.07} ($\pm$0.07)  &95.03  &98.20  &98.08  &99.04  \\ \hline
\end{tabular}
\end{table*}

\begin{table*}[tbp]
\small
\centering
\caption{ Comparison of the mean test accuracy (\%) and standard
deviation on Fashion-MNIST under weak, middle and strong Supervision
degree, where the best performance is highlighted in bold.}
\label{tab:fashion_compare}
\begin{tabular}{|c|l|c|cc|cc|}
\hline
\multicolumn{1}{|l|}{Supervision} &
  N\_c &
  LeNet-5 &
  HOG-I &
  HOG-II &
  IPHop-I &
  IPHop-II \\ \hline
\multirow{4}{*}{Weak}   & 1    &41.18 ($\pm$5.06)  &\textbf{49.80} ($\pm$ 4.29) &10.00 ($\pm$ 0.00) &46.73 ($\pm$4.87)  &10.00 ($\pm$0.00)  \\
                        & 2    &50.65 ($\pm$5.36)  &54.43 ($\pm$ 4.42) &39.85 ($\pm$ 2.07) &\textbf{56.57} ($\pm$2.16)  &47.17 ($\pm$2.42)  \\
                        & 4    &56.22 ($\pm$4.23)  &\textbf{60.42} ($\pm$ 1.99) &41.53 ($\pm$ 2.21) &59.21 ($\pm$3.39)  &52.48 ($\pm$2.73)  \\
                        & 8    &60.54 ($\pm$3.85)  &\textbf{64.25} ($\pm$ 1.58) &54.29 ($\pm$ 2.28) &62.90 ($\pm$1.91)  &65.44 ($\pm$1.10)  \\ \hline
\multirow{4}{*}{Middle} & 16   &61.34 ($\pm$3.17)  &68.22 ($\pm$ 1.71) &66.38 ($\pm$ 1.33) &69.37 ($\pm$1.47)  &\textbf{73.93} ($\pm$1.41)  \\
                        & 32   &67.49 ($\pm$2.88)  &71.60 ($\pm$ 0.75) &73.02 ($\pm$ 0.74) &71.47 ($\pm$0.89)  &\textbf{77.86} ($\pm$0.88) \\
                        & 64   &71.58 ($\pm$2.33)  &73.33 ($\pm$ 0.48) &77.60 ($\pm$ 0.66) &74.44 ($\pm$0.52)  &\textbf{80.88} ($\pm$0.89)  \\
                        & 128  &75.04 ($\pm$1.65)  &75.49 ($\pm$ 0.68) &80.12 ($\pm$ 0.57) &76.81 ($\pm$0.28)  &\textbf{83.54} ($\pm$0.37)  \\ \hline
\multirow{6}{*}{Strong} & 256  &76.81 ($\pm$3.05)  &77.57 ($\pm$ 0.52) &82.78 ($\pm$ 0.41) &78.74 ($\pm$0.37)  &\textbf{85.59} ($\pm$0.33)  \\
                        & 512  &82.38 ($\pm$2.19)  &79.05 ($\pm$ 0.37) &84.19 ($\pm$ 0.17) &80.29 ($\pm$0.31)  &\textbf{87.41} ($\pm$0.25)  \\
                        & 1024 &84.51 ($\pm$2.34)  &80.56 ($\pm$ 0.38) &86.00 ($\pm$ 0.25) &81.99 ($\pm$0.26)  &\textbf{88.81} ($\pm$0.20)  \\
                        & 2048 &87.13 ($\pm$2.06)  &81.91 ($\pm$ 0.24) &87.14 ($\pm$ 0.17) &83.59 ($\pm$0.26)  &\textbf{89.93} ($\pm$0.13)  \\
                        & 4096 &88.97 ($\pm$0.56)  &83.06 ($\pm$ 0.18) &88.35 ($\pm$ 0.08) &85.01 ($\pm$0.11)  &\textbf{91.03} ($\pm$0.16)  \\
                        & Full &89.54 ($\pm$0.33)  &83.52  &88.84  &85.77  &\textbf{91.37}  \\ \hline
\end{tabular}
\end{table*}

\begin{figure*}[tbp]
\centering
\begin{subfigure}{0.45\textwidth}
\centering
    \includegraphics[width=0.95\linewidth]{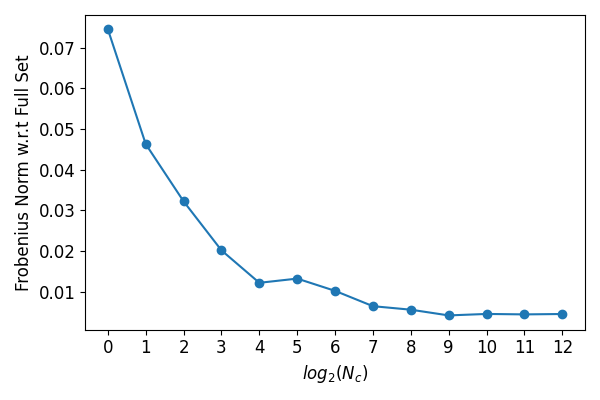}
    \caption{Local Saab filters in Hop-1}
\end{subfigure}
\begin{subfigure}{0.45\textwidth}
\centering
    \includegraphics[width=0.95\linewidth]{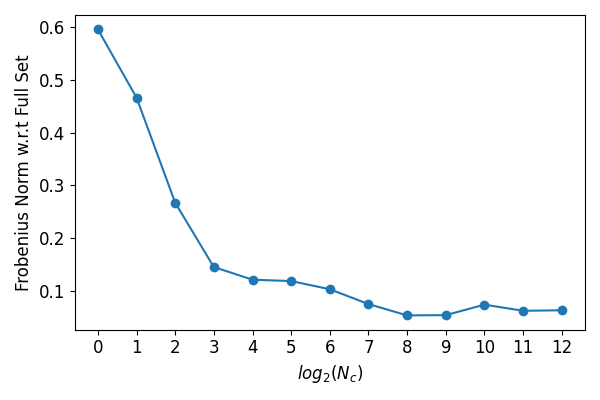}
    \caption{Local Saab filters in Hop-2}
\end{subfigure}\\
\begin{subfigure}{0.45\textwidth}
\centering
    \includegraphics[width=0.95\linewidth]{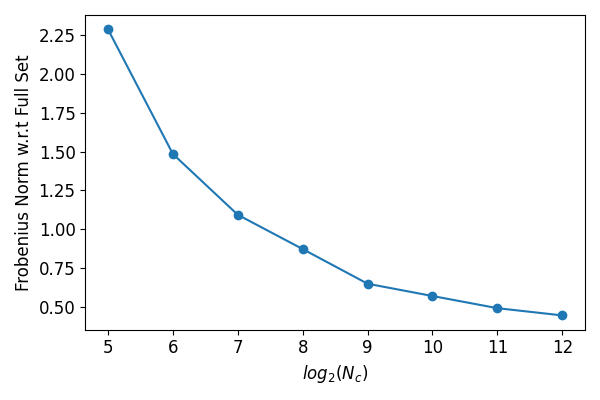}
    \caption{Global Saab filters in Hop-1}
\end{subfigure}
\begin{subfigure}{0.45\textwidth}
\centering
    \includegraphics[width=0.95\linewidth]{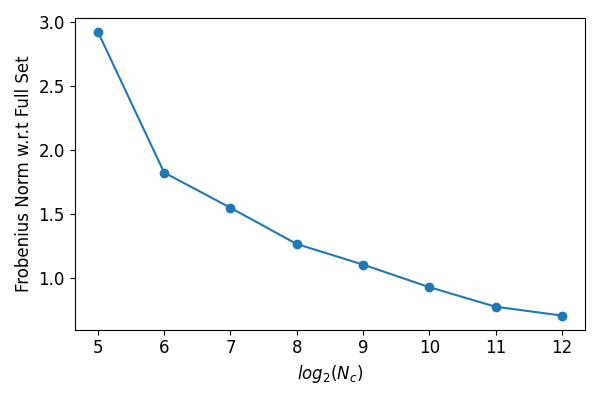}
    \caption{Global Saab filters in Hop-2}
\end{subfigure}
\caption{The plot of Frobenius norms of difference matrices between the
covariance matrices learned under different supervision levels and the
one learned from the full set for MNIST.} \label{fig:fro_1}
\end{figure*}

\section{Discussion} \label{sec:discussion}

The superiority of hybrid IPHop under both weak and strong supervision
conditions is clearly demonstrated in Fig. \ref{fig:hybrid_compare}. 
We would like to provide some explanations in this section.

\begin{figure*}[tbp]
\centering
\begin{subfigure}{0.48\textwidth}
\centering
    \includegraphics[width=1.0\linewidth]{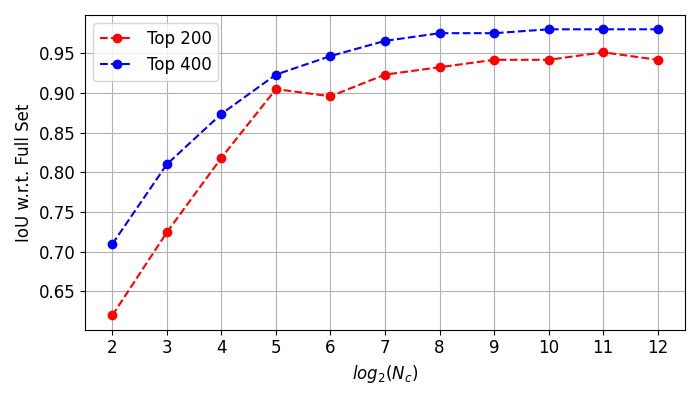}
    \caption{MNIST}
    \label{fig:dft_1a}
\end{subfigure}
\begin{subfigure}{0.48\textwidth}
\centering
    \includegraphics[width=1.0\linewidth]{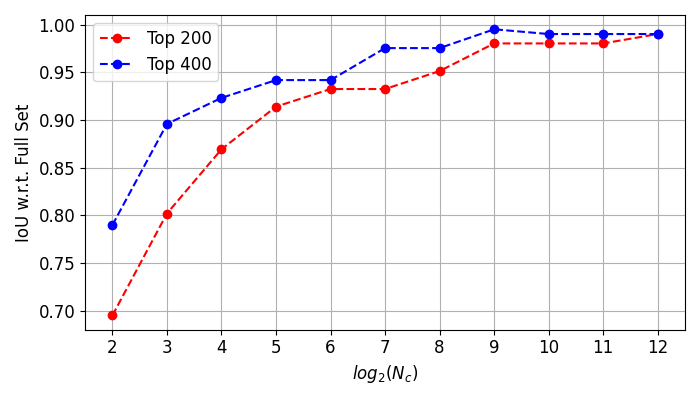}
    \caption{Fashion-MNIST}
    \label{fig:dft_1b}
\end{subfigure}
\caption{IoU scores between the feature sets selected using full
training size and using $N_c$ on MNIST and Fashion-MNIST.}
\label{fig:dft_1}
\end{figure*}

\begin{itemize}
\item Robustness in Representation Learning \\
The IPHop representation is determined by Saab filters.  Saab filters are
obtained by PCA, which is an eigen-analysis of the covariance matrix of
input vectors. If the covariance matrix converges fast as the training
sample number increases, then IPHop's feature learning is robust with
respect to supervision degree. We show the Frobenius norm of the
difference matrix between the covariance matrix derived by $N_c$
training images and the full training size in Fig.~\ref{fig:fro_1}.
There are four cases; namely, the local and global Saab filters in Hop-1
and Hop-2, respectively. The results are averaged among 5 runs.  We see
that the Forbenius norm of the difference covariance matrices is already
small even for $N_c=1$. This is because one image contains many small
patches which contribute to a robust covariance matrix. 
\item Robustness in Feature Learning \\
To demonstrate the robustness of DFT, we measure the overlapping of the
selected feature set based on $N_c$ training samples and that based on
the full training size ($N_c=6K$). These two sets are denoted by
$\{F\}_{N_c}$ and $\{F\}_{full}$, respectively.  We define an
intersection-over-union (IoU) score as
\begin{equation}\label{eq:iou}
IoU_{N_c} = \frac{\left | \{F\}_{full} \bigcap \{F\}_{N_c} 
\right |}{\left | \{F\}_{full} \bigcup \{F\}_{N_c} \right |},
\end{equation}
where the numerator represents for the number of features agreed between
the two subsets while the denominator represents the number of features
selected by at least one of the two subsets.  For each $N_c$, there
exists randomness in selecting a subsets of labeled samples. To
eliminate the randomness, we calculate the averaged IoU values with 10
runs.  The IoU values of selecting the top 200-D and 400-D from the
1024-D HOG features for MNIST and Fashion-MNIST, respectively, are shown
in Fig.~\ref{fig:dft_1}.  We see that, as the number of labeled data
increases, the IoU score increases. With a small $N_c$ value (say, 32)
the IoU score can already reach 90\%.  It clearly shows that DFT
is a semi-supervised feature selection tool and it can work well under
very weak supervision condition. 
\item Robustness in Decision Learning \\ 
The KNN classifier is used when the training number is small. It is an
exemplar-based classifier.  Instead of minimizing the loss using labeled
data, it finds the most similar training sample based on the Euclidean
distance in the feature space. However, it cannot capture the data
manifold that often lies in a higher dimensional feature space.  As the
number of labeled data increases, XGBoost is more powerful. It minimizes
the cross-entropy loss with a gradient boosting technique. XGBoost is a
decision tool based on ensemble learning which explains its robust
decision behavior.
\end{itemize}

\begin{figure*}[tbp]
\centerline{\includegraphics[width=0.6\linewidth]{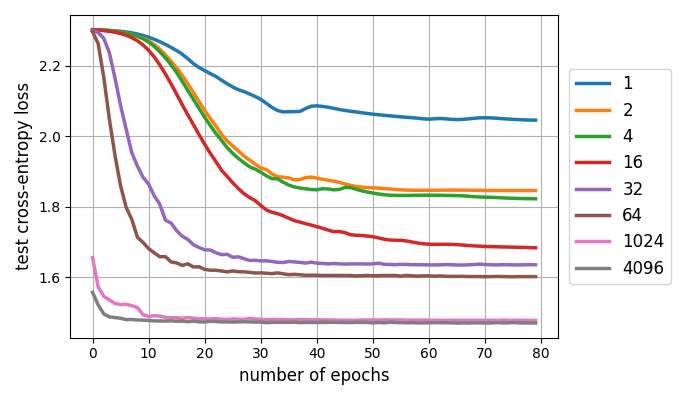}}
\caption{Learning curve using LeNet-5 on MNIST dataset with selected supervision levels.}
\label{fig:lenet_learning_curve}
\end{figure*}

On the other hand, it is desired to understand the behaviour of LeNet-5
under different supervision levels.  We show the learning curves of
LeNet-5 on MNIST using $N_c=2^{n_c}$, $n_c=0, 1, \cdots, 12$, in
Fig.~\ref{fig:lenet_learning_curve}, which are expressed as the
cross-entropy losss as a function of the epoch number.  The batch size
in each epoch is set to the total number of labeled images if $N_c \le
16$ and 128 if $N_c>16$.  The loss curves are averaged among 5 random
runs.  The loss decreases slowly and converges at a higher loss value
when $N_c$ is small. In contrast, it decreases faster and converges at a
lower loss value when $N_c$ is larger. Clearly, the learning performance
of LeNet-5 in both convergence rates and converged loss values is highly
dependent on the number of the labeled samples. 

\section{Conclusion and Future Work}\label{sec:conclusion}

In this work, we compared the supervision-scalability of three learning
systems; namely, the HOG-based and IPHop-based learning systems and
LeNet-5, which is an representative deep-learning system.  Both
HOG-based and IPHop-based learning systems work better than LeNet-5
under weak supervision. As the supervision degree goes higher, the
performance gap narrows. Yet, IPHop-II still outperforms LeNet-5 on
Fashion-MNIST under strong supervision. 

It is well known that it is essential to have a sufficient amount of
labeled data for deep learning systems to work properly. Data
augmentation and adoption of pre-trained networks are two commonly used
techniques to overcome the problem of insufficient training data at the
cost of larger model sizes and higher computational cost.  Our
performance benchmarking study is only preliminary. In the future, we
would like to conduct further investigation by considering supervision
scalability, tradeoff of accuracy, model sizes and computational
complexity jointly. 

\section*{Acknowledgement}\label{sec:acknowledgement}

The authors acknowledge the Center for Advanced Research Computing
(CARC) at the University of Southern California for providing computing
resources that have contributed to the research results reported within
this publication.

\bibliographystyle{ieeetr}
\bibliography{refs}

\end{document}